\documentclass{article}

\usepackage{booktabs,siunitx}

\usepackage{times}
\usepackage{graphicx} 
\usepackage{subfig}

\usepackage{natbib}

\usepackage{algorithm}
\usepackage{algorithmic}

\usepackage{hyperref}

\usepackage[accepted]{icml2017}
\usepackage{amsmath}
\usepackage{amsfonts}

\icmltitlerunning{Schema Networks: Zero-shot Transfer with a Generative Causal Model of Intuitive Physics}

\begin{document}

\twocolumn[
\icmltitle{Schema Networks: Zero-shot Transfer with a Generative Causal Model of Intuitive Physics}




\begin{icmlauthorlist}
\icmlauthor{Ken Kansky}{}
\icmlauthor{Tom Silver}{}
\icmlauthor{David A. M\'ely}{}
\icmlauthor{Mohamed Eldawy}{}
\icmlauthor{Miguel L\'azaro-Gredilla}{}
\icmlauthor{Xinghua Lou}{}
\icmlauthor{Nimrod Dorfman}{}
\icmlauthor{Szymon Sidor}{}
\icmlauthor{Scott Phoenix}{}
\icmlauthor{Dileep George}{}
\end{icmlauthorlist}


\icmlcorrespondingauthor{Ken Kansky}{ken@vicarious.com}
\icmlcorrespondingauthor{Tom Silver}{tom@vicarious.com}

\icmlkeywords{schema networks, intuitive physics, object-oriented, markov decision process, factor graph}

\vskip 0.3in
]



\printAffiliationsAndNotice{\vskip 0.1in
All authors affiliated with Vicarious AI, California, USA}  

\begin{abstract}
The recent adaptation of deep neural network-based methods to reinforcement learning and planning domains has yielded remarkable progress on individual tasks. Nonetheless, progress on task-to-task transfer remains limited. In pursuit of efficient and robust generalization, we introduce the Schema Network, an object-oriented generative physics simulator capable of disentangling multiple causes of events and reasoning backward through causes to achieve goals. The richly structured architecture of the Schema Network can learn the dynamics of an environment directly from data. We compare Schema Networks with Asynchronous Advantage Actor-Critic and Progressive Networks on a suite of Breakout variations, reporting results on training efficiency and zero-shot generalization, consistently demonstrating faster, more robust learning and better transfer. We argue that generalizing from limited data and learning causal relationships are essential abilities on the path toward generally intelligent systems.
\end{abstract}

\section{Introduction}
\label{sec:introduction}

\begin{figure}[t]
\vskip 0.2in
\begin{center}
\centerline{\includegraphics[width=\columnwidth]{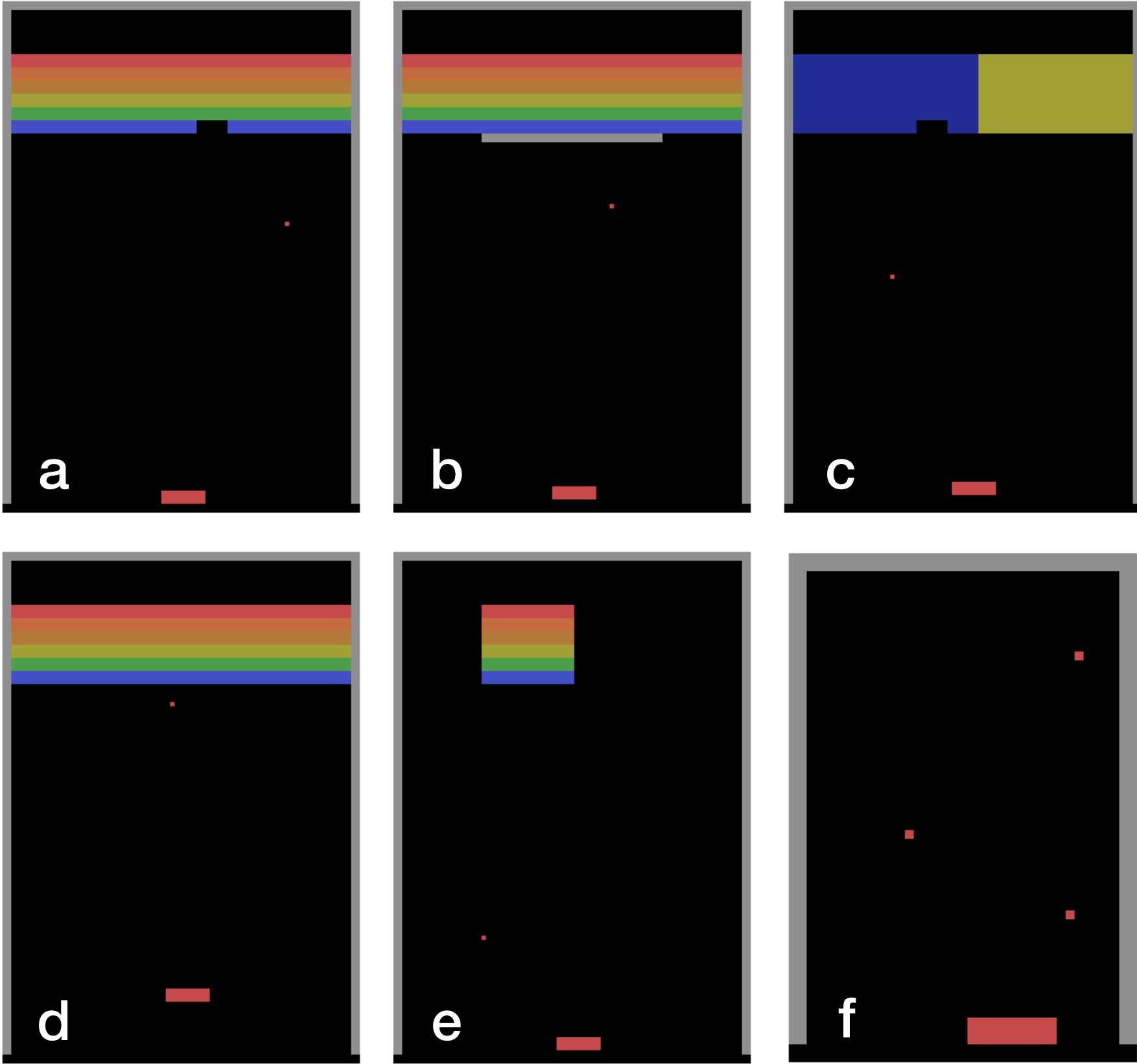}}
\caption{Variations of Breakout. From top left: standard version, middle wall, half negative bricks, offset paddle, random target, and juggling. After training on the standard version, Schema Networks are able to generalize to the other variations without any additional training.}
\label{fig:breakout_variations}
\end{center}
\vskip -0.2in
\end{figure}

A longstanding ambition of research in artificial intelligence is to efficiently generalize experience in one scenario to other similar scenarios. Such generalization is essential for an embodied agent working to accomplish a variety of goals in a changing world. Despite remarkable progress on individual tasks like Atari 2600 games \cite{mnih2015human, van2016deep, mnih2016asynchronous} and Go \cite{silver2016mastering}, the ability of state-of-the-art models to \textit{transfer} learning from one environment to the next remains limited. For instance, consider the variations of Breakout illustrated in Fig.~\ref{fig:breakout_variations}. In these environments the positions of objects are perturbed, but the object movements and sources of reward remain the same. While humans have no trouble generalizing experience from the basic Breakout to its variations, deep neural network-based models are easily fooled \cite{taylor2009transfer, rusu2016progressive}.

The model-free approach of deep reinforcement learning (Deep RL) such as the Deep-Q Network and its descendants is inherently hindered by the same feature that makes it desirable for single-scenario tasks: it makes no assumptions about the structure of the domain. Recent work has suggested how to overcome this deficiency by utilizing \textit{object-based representations} \cite{diuk2008object, usunier2016episodic}. Such a representation is motivated by the well-acknowledged Gestalt principle, which states that the ability to perceive objects as a bounded figure in front of an unbounded background is fundamental to all perception \cite{weiten2012psychology}. \citet{battaglia2016interaction} and \citet{chang2016compositional} go further, learning relations to predict object interactions.

While object-based and relational representations have shown great promise alone, they stop short of modeling \textit{causality} -- the ability to reason about previous observations and explain away alternative causes. A causal model is essential for regression planning, in which an agent works backward from a desired future state to produce a plan \cite{anderson1990cognitive}. Reasoning backward and allowing for multiple causation requires a framework like Probabilistic Graphical Models (PGMs), which can natively support explaining away \cite{koller2009probabilistic}.

Here we introduce Schema Networks -- a generative model for object-oriented reinforcement learning and planning\footnote{We borrow the term ``schema'' from \citet{drescher1991made}, whose schema mechanism inspired the early development of our model.}. Schema Networks incorporate key desiderata for the flexible and compositional transfer of learned prior knowledge to new settings. 1) Knowledge is represented with ``schemas'' -- local cause-effect relationships involving one or more object entities; 2) In a new setting, these cause-effect relationships are traversed to guide action selection; and 3) The representation deals with uncertainty, multiple-causation, and explaining away in a principled way. We first describe the representational framework and learning algorithms and then demonstrate how action policies can be generated by treating planning as inference in a factor graph. We evaluate the end-to-end system on Breakout variations and compare against Asynchronous Advantage Actor-Critic (A3C) \cite{mnih2016asynchronous} and Progressive Networks (PNs) \cite{rusu2016progressive}, the latter of which extends A3C explicitly to handle transfer. We show that the structure of the Schema Network enables efficient and robust generalization beyond these Deep RL models.

\section{Related Work}
\label{sec:relatedwork}

The field of reinforcement learning has witnessed significant progress with the recent adaptation of deep learning methods to traditional frameworks like Q-learning. Since the introduction of the Deep Q-network (DQN) \cite{mnih2015human}, which uses experience replay to achieve human-level performance on a set of Atari 2600 games, several innovations have enabled faster convergence and better performance with less memory. The asynchronous methods introduced by \citet{mnih2016asynchronous} exploit multiple agents acting in copies of the same environment, combining their experiences into one model. As the Asynchronous Advantage Actor-Critic (A3C) is the best among these methods, we use it as our primary comparison.

Model-free Deep RL models like A3C are unable to substantially generalize beyond their training experience \cite{jaderberg2016reinforcement, rusu2016progressive}. To address this limitation, recent work has attempted to introduce more structure into neural network-based models. The Interaction Network \cite{battaglia2016interaction} (IN) and the Neural Physics Engine (NPE) \cite{chang2016compositional} use object-level and pairwise relational representations to learn models of intuitive physics. The primary advantage of these models is their amenability to gradient-based methods, though such techniques might be applied to Schema Networks as well. Schema Networks offer two key advantages: latent physical properties and relations need not be hardcoded, and planning can make use of backward search, since the model can distinguish different causes. Furthermore, neither INs nor NPEs have been applied in RL domains. Progress in model-based Deep RL has thus far been limited, though methods like Embed to Control \cite{watter2015embed}, Value Iteration Networks \cite{tamar2016value}, and the Predictron \cite{silver2016predictron} demonstrate the promise of this direction. However, these approaches do not exploit the object-relational representation of INs or NPEs, nor do they incorporate a backward model for regression planning.

Schema Networks build upon the ideas of the Object-Oriented Markov Decision Process (OO-MDP) introduced by \citet{diuk2008object} (see also \citet{scholz2014physics}). Related frameworks include relational and first-order logical MDPs \cite{guestrin2003generalizing}. These formalisms, which harken back to classical AI's roots in symbolic reasoning, are designed to enable robust generalization. Recent work by \citet{garnelo2016towards} on ``deep symbolic reinforcement learning'' makes this connection explicit, marrying first-order logic with deep RL. This effort is similar in spirit to our work with Schema Networks, but like INs and NPEs, it lacks a mechanism to learn disentangled causes of the same effect and cannot perform regression planning.

Schema Networks transfer experience from one scenario to other similar scenarios that exhibit repeatable structure and sub-structure \cite{taylor2009transfer}. \citet{rusu2016progressive} show how A3C can be augmented to similarly exploit common structure between tasks via Progressive Networks (PNs). A PN is constructed by successively training copies of A3C on each task of interest. With each new task, the existing network is frozen, another copy of A3C is added, and lateral connections between the frozen network and the new copy are established to facilitate transfer of features learned during previous tasks. One obvious limitation of PNs is that the number of network parameters must grow quadratically with the number of tasks. However, even if this growth rate was improved, the PN would still be unable to generalize from biased training data without continuing to learn on the test environment. In contrast, Schema Networks exhibit zero-shot transfer.

Schema Networks are implemented as probabilistic graphical models (PGMs), which provide practical inference and structure learning techniques. Additionally, inference with uncertainty and explaining away are naturally supported by PGMs. We direct the readers to \cite{koller2009probabilistic} and \cite{jordan1998learning} for a thorough overview of PGMs. In particular, early work on factored MDPs has demonstrated how PGMs can be applied in RL and planning settings \cite{guestrin2003efficient}.

\section{Schema Networks}

\subsection{MDPs and Notation}
The traditional formalism for the Reinforcement Learning problem is the Markov Decision Process (MDP). An MDP $M$ is a five-tuple $(\mathcal{S}, \mathcal{A}, T, R, \gamma)$, where $\mathcal{S}$ is a set of states, $\mathcal{A}$ is a set of actions, $T(s^{(t+1)} | s^{(t)}, a^{(t)})$ is the probability of transitioning from state $s^{(t)} \in \mathcal{S}$ to $s^{(t+1)} \in \mathcal{S}$ after action $a^{(t)} \in \mathcal{A}$, $R(r^{(t+1)} | s^{(t)}, a^{(t)})$ is the probability of receiving reward $r^{(t+1)} \in \mathbb{R}$ after executing action $a^{(t)}$ while in state $s^{(t)}$, and $\gamma \in [0, 1]$ is the rate at which future rewards are exponentially discounted.

\begin{figure}[tb]
\begin{center}
\centerline{\includegraphics[width=\columnwidth]{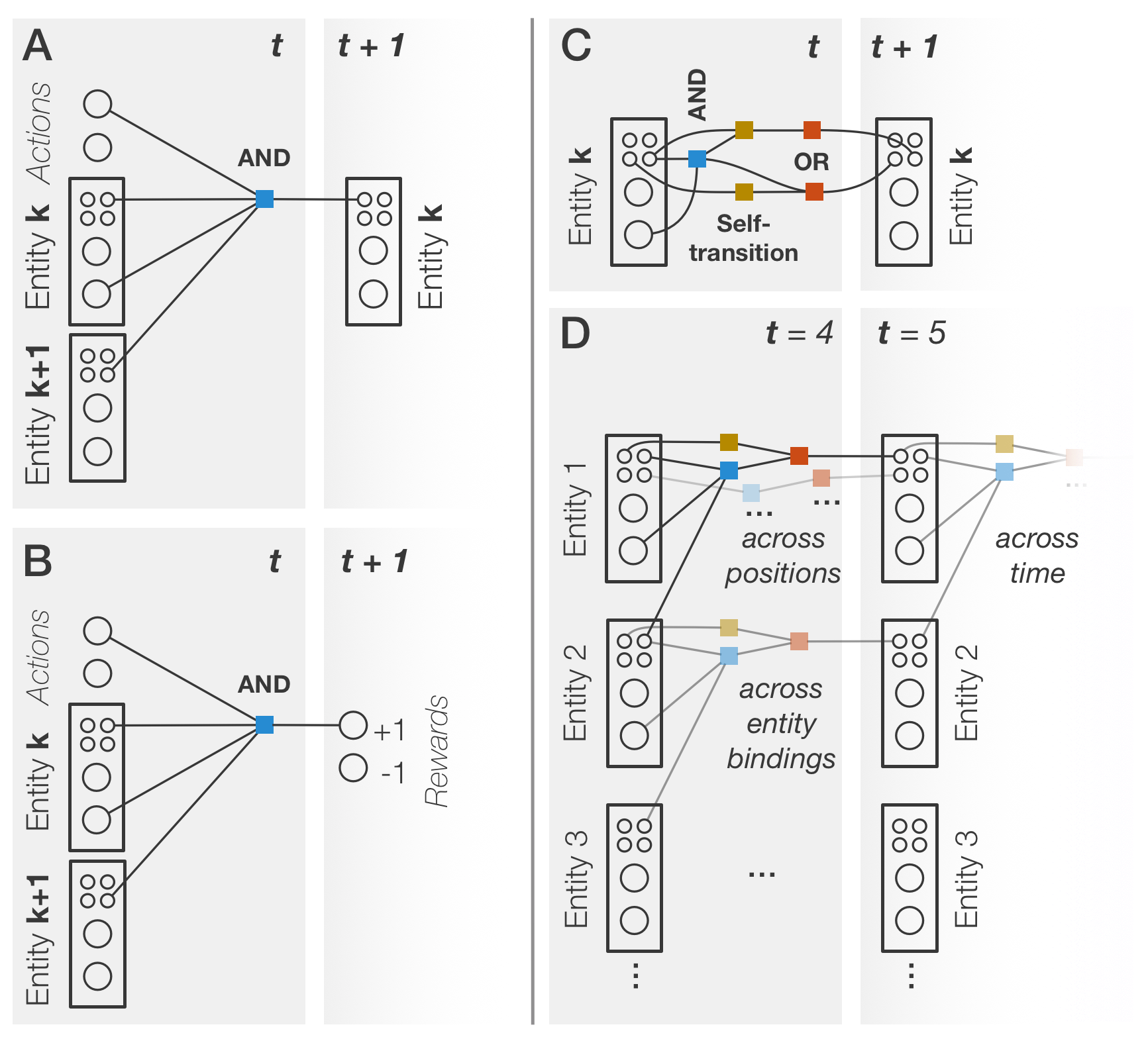}}
\caption{\textbf{Architecture of a Schema Network.} An ungrounded schema is a template for a factor that predicts either the value of an entity-attribute (A) or a future reward (B) based on entity states and actions taken in the present. Self-transitions (C) predict that entity-attributes remain in the same state when no schema is active to predict a change. Self-transitions allow continuous or categorical variables to be represented by a set of binary variables (depicted as smaller nodes). The grounded schema factors, instantiated from ungrounded schemas at all positions, times, and entity bindings, are combined with self-transitions to create a Schema Network (D).}
\label{fig:schema-network-example}
\end{center}
\end{figure}

\subsection{Model Definition}
A Schema Network is a structured generative model of an MDP. We first describe the architecture of the model informally. An image input is parsed into a list of \textit{entities}, which may be thought of as instances of objects in the sense of OO-MDPs \cite{diuk2008object}. All entities share the same collection of \textit{attributes}. We refer to a specific attribute of a specific entity as an \textit{entity-attribute}, which is represented as a binary variable to indicate the presence of that attribute for an entity. An \textit{entity state} is an assignment of states to all attributes of the entity, and the complete model state is the set of all entity states.

A \textit{grounded schema} is a binary variable associated with a particular entity-attribute in the next timestep, whose value depends on the present values of a set of binary entity-attributes.
The event that one of these present entity-attributes assumes the value 1 is called a \textit{precondition} of the grounded schema.
When all preconditions of a grounded schema are satisfied, we say that the schema is active, and it predicts the activation of its associated entity-attribute. Grounded schemas may also predict \textit{rewards} and may be conditioned on \textit{actions}, both of which are represented as binary variables.
For instance, a grounded schema might define a distribution over Entity 1's ``position'' attribute at time 5, conditioned on Entity 2's ``position'' attribute at time 4 and the action ``UP'' at time 4. Grounded schemas are instantiated from \textit{ungrounded schemas}, which behave like templates for grounded schemas to be instantiated at different times and in different combinations of entities. For example, an ungrounded schema could predict the ``position'' attribute of Entity $x$ at time $t+1$ conditioned on the ``position'' of Entity $y$ at time $t$ and the action ``UP'' at time $t$; this ungrounded schema could be instantiated at time $t=4$ with $x = 1$ and $y = 2$ to create the grounded schema described above. In the case of attributes like ``position'' that are inherently continuous or categorical, several binary variables may be used to discretely approximate the distribution (see the smaller nodes in Figure \ref{fig:schema-network-example}). A \textbf{\textit{Schema Network}} is a factor graph that contains all grounded instantiations of a set of ungrounded schemas over some window of time, illustrated in Figure \ref{fig:schema-network-example}.

We now formalize the Schema Network factor graph. For simplicity, suppose the number of entities and the number of attributes are fixed at $N$ and $M$ respectively. Let $E_i$ refer to the $i^{th}$ entity and let $\alpha_{i,j}^{(t)}$ refer to the $j^{th}$ attribute value of the $i^{th}$ entity at time $t$. We use the notation $E_i^{(t)} = (\alpha_{i,1}^{(t)}, ..., \alpha_{i,M}^{(t)})$ to refer to the state of the $i^{th}$ entity at time $t$. The complete state of the MDP modeled by the network at time $t$ is then $s^{(t)} = (E_1^{(t)}, ..., E_N^{(t)})$. Actions and rewards are also represented with sets of binary variables, denoted $a^{(t)}$ and $r^{(t+1)}$ respectively. A Schema Network for time $t$ will contain the variables in $s^{(t)}$, $a^{(t)}$, $s^{(t+1)}$, and $r^{(t+1)}$.

Let $\phi^{k}$ denote the variable for grounded schema $k$. $\phi^{k}$ is bound to a specific entity-attribute $\alpha_{i,j}$ and activates it when the schema is active. Multiple grounded schemas can predict the same attribute, and these predictions are combined through an OR gate. For binary variables $v_1, ..., v_n$, let $\text{AND}(v_1, ..., v_n) = \prod_{i=1}^n P(v_i = 1)$, and $\text{OR}(v_1, ..., v_n) =1 - \prod_{i=1}^n (1-P(v_i = 1))$. A grounded schema is connected to its precondition entity-attributes with an AND factor, written as $\phi^{k} = \text{AND}(\alpha_{i_1,j_1}, ..., \alpha_{i_H,j_H}, a)$ for $H$ entity-attribute preconditions and an optional action $a$. There is no restriction on how many entities or attributes from a single entity can be preconditions of a grounded schema.

An ungrounded schema (or template) is represented as  $\Phi_{l}(E_{x_1}, ..., E_{x_H}) = \text{AND}(\alpha_{x_1,y_1}, \alpha_{x_1,y_2}..., \alpha_{x_H,y_H})$, where $x_h$ determines the relative entity index of the $h$-th precondition and $y_h$ determines which attribute variable is the precondition. The ungrounded schema is a template that can be bound to multiple specific entities and locations to generate grounded schemas.

A subset of attributes corresponds to discrete positions. These attributes are treated differently from all others, whose semantic meanings are unknown to the model. When a schema predicts a movement to a new position, we must inform the previously active position attribute to be inactive unless there is another schema that predicts it to remain active. We introduce a \textit{self-transition} variable to represent the probability that a position attribute will remain active in the next time step when no schema predicts a change from that position. We compute the self-transition variable as $\Lambda_{i,j} = \text{AND}(\neg \phi^1, ..., \neg \phi^k, s_{i,j})$ for entity $i$ and position attribute $j$, where the set $\phi^1 ... \phi^k$ includes all schemas that predict the future position of the same entity $i$ and include $s_{i,j}$ as a precondition.

With these terms defined, we may now compute the transition function, which can be factorized as $T(s^{(t+1)} | s^{(t)}, a^{(t)}) = \prod_{i=1}^N \prod_{j = 1}^M T_{i, j}(s_{i, j}^{(t+1)} | s^{(t)}, a^{(t)})$. An entity-attribute is active at the next time step if either a schema predicts it to be active or if its self-transition variable is active: $T_{i, j}(s_{i,j}^{(t+1)} | s^{(t)}) = \text{OR}(\phi^{k_1}, ..., \phi^{k_Q}, \Lambda_{i,j})$, where $k_1 ... k_Q$ are the indices of all grounded schemas that predict $s_{i,j}$.

\subsection{Construction of Entities and Attributes}

In practice we assume that a vision system is responsible for detecting and tracking entities in an image. It is therefore largely up to the vision system to determine what constitutes an entity. Essentially any trackable image feature could be an entity, which most typically includes objects, their boundaries, and their surfaces. Recent work has demonstrated one possible method for unsupervised entity construction using autoencoders \cite{garnelo2016towards}. Depending on the task, Schema Networks could learn to reason flexibly at different levels of representation. For example, using entities from surfaces might be most relevant for predicting collisions, while using one entity per object might be most relevant for predicting whether it can be controlled by an action. The experiments in this paper utilize surface entities, described further in Section \ref{experiments}.

Similarly, entity attributes can be provided by the vision system, and these attributes typically include: color/appearance, surface/edge orientation, object category, or part-of an object category (e.g. front-left tire). For simplicity we here restrict the entities to have fully observable attributes, but in general they could have latent attributes such as ``bounciness'' or ``magnetism.''

\subsection{Connections to Existing Models}

Schema Networks are closely related to Object-Oriented MDPs (OO-MDPs) \cite{diuk2008object} and Relational MDPs (R-MDPs) \cite{guestrin2003generalizing}. However, neither OO-MDPs nor R-MDPs define a transition function with an explicit OR of possible causes, and traditionally transition functions have not been learned in these models. In contrast, Schema Networks provide an explicit OR to reason about multiple causation, which enables regression planning. Additionally, the structure of Schema Networks is amenable to efficient learning.

Schema Networks are also related to the recently proposed Interaction Network (IN) \cite{battaglia2016interaction} and Neural Physics Engine (NPE) \cite{chang2016compositional}. At a high level, INs, NPEs, and Schema Networks are much alike -- objects are to entities as relations are to schemas. However, neither INs nor NPEs are generative and hence do not support regression planning from a goal through causal chains. Because Schema Networks are generative models, they support more flexible inference and search strategies for planning. Additionally, the learned structures in Schema Networks are amenable to human interpretation, explicitly factorizing different causes, making prediction errors easier to relate to the learned model parameters.


\section{Learning and Planning in Schema Networks}

In this section we describe how to train Schema Networks (i.e., learn its structure) from interactions with an environment, as well as how they can be used to perform planning. Planning is not only necessary at test time to maximize reward, but also can be used to improve exploration during the training procedure.

\subsection{Training Procedure}
\label{sec:schema-learning}

Given a series of actions, rewards and images, we represent each possible action and reward with a binary variable, and we convert each image into a set of entity states. The number of entities is allowed to vary between adjacent frames, accounting for objects appearing or moving out of view.

Given a dataset of entity states over time, we preprocess the entity states into a representation that is more convenient for learning. For $N$ entities observed over $T$ timesteps, we wish to predict $\alpha_{i, j}^{(t)}$ on the basis of the attribute values of the $i^{th}$ entity and its spatial neighbors at time $t-1$ (for $1 \le i \le N$ and $2 \le t \le T$). The attribute values of $E_i^{(t-1)}$ and its neighbors can be represented as a row vector of length $MR$, where $M$ is the number of attributes and $R-1$ is the number of neighbor positions of each entity, determined by a fixed radius. Let $X \in \{0, 1\}^{D \times D'}$ be the arrangement of all such vectors into a binary matrix, with $D = NT$ and $D' = MR$. Let $y \in \{0,1\}^{D}$ be a binary vector such that if row $r$ in $X$ refers to $E_i^{(t-1)}$, then $y_r=\alpha_{i, j}^{(t)}$. Schemas are then learned to predict $y$ from $X$ using the method described in Section \ref{sec:schemalearning}.

While gathering data, actions are chosen by planning using the schemas that have been learned so far. This planning algorithm is described in Section \ref{alg:planning}. We use an $\varepsilon$-greedy approach to encourage exploration, taking a random action at each timestep with small probability. We found no need to perform any additional policy learning, and after convergence predictions were accurate enough to allow for successful planning. As shown in Section \ref{experiments}, since learning only involves understanding the dynamics of the game, transfer learning is simplified and there is no need for policy adaptation.

\subsection{Schema Learning}
\label{sec:schemalearning}




Structure learning in graphical models is a well studied topic in machine learning \cite{koller2009probabilistic, jordan1998learning}. To learn the structure of the Schema Network, we cast the problem as a supervised learning problem over a discrete space of parameterizations (the schemas), and then apply a greedy algorithm that solves a sequence of LP relaxations. See \citet{jaakkola2010learning} for further work on applying LP relaxations to structure learning.

With $X$ and $y$ defined above, the learning problem is to find a mapping

$$y = f_W(X) = \overline {\overline{X}W}\vec{1}
$$

where all the involved variables are binary and operations follow Boolean logic: addition corresponds to ORing, and overlining to negation. $W\in \{0,1\}^{D' \times L}$ is a binary matrix, with each column representing one (ungrounded) schema for at most $L$ schemas. The elements set to 1 in each schema represent an existing connection between that schema and an input condition (see Fig. \ref{fig:schema-network-example}). The outputs of each individual schema are ORed to produce the final prediction.

We would like to minimize the prediction error of Schema Networks while keeping them as simple as possible. A suitable objective function is
\begin{equation}
\label{eq:schemalearningobj}
\min_{W\in \{0,1\}^{D' \times L}} \frac{1}{D}|y - f_W(X)|_1 + C|W|_1,
\end{equation}
where the first term computes the prediction error, the second term estimates the complexity and parameter $C$ controls the trade-off between both. This is an NP-hard problem for which we cannot hope to find an exact solution, except for very small environments.

We consider a greedy solution in which linear programming (LP) relaxations are used to find each new schema. Starting from the empty set, we greedily add schemas (columns to $W$) that have perfect precision and increase recall for the prediction of $y$ (See Algorithm~1 in the Supplementary). In each successive iteration, only the input-output pairs for which the current schema network is predicting an output of zero are passed. This procedure monotonically decreases the prediction error of the overall schema network, while increasing its complexity. The process stops when we hit some predefined complexity limit. In our implementation, the greedy schema selection produces very sparse schemas, and we simply set a limit to the number of schemas to add. For this algorithm to work, no contradictions can exist in the input data (such as the same input appearing twice with different labels). Such contradictions might appear in stochastic environments and would not be artifacts in real environments, so we preprocess the input data to remove them.

\subsection{Planning as Probabilistic Inference}
\label{alg:planning}

The full Schema Network graph (Fig. \ref{fig:schema-network-example}) provides a probabilistic model for the set of rewards that will be achieved by a sequence of actions. Finding the sequence of actions that will result in a given set of rewards becomes then a MAP inference problem. This problem can be addressed approximately using max-product belief propagation (MPBP) \cite{attias2003planning}. Another option is variational inference. \citet{cheng2013variational} use variational inference for planning but resort to MPBP to optimize the variational free energy functional. We will follow the first approach.

Without loss of generality, we will consider the present time step to be $t=0$. The state, action and reward variables for $t\leq0$ are observed, and we will consider inference over the unobserved variables in a look-ahead window of size\footnote{In contrast with MDPs, the reward is discounted with a rolling square window instead of an exponentially weighted one.} $T$, $\{s^{(t)}, a^{(t)}, r^{(t)}\}_{t=0}^{T-1}$. Since the Schema Network is built exclusively of \emph{compatibility factors} that can take values 0 or 1, any variable assignment is either impossible or equally probable under the joint distribution of the graph. Thus, if we want to know if there exists any global assignment that activates a binary variable (say, variable $r_{(+)}^{(t)}$ signaling positive reward at some future time $t>0$), we should look at the max-marginal $\tilde p(r_{(+)}^{(t)}=1)$. It will be 0 if no global assignment compatible with both the SN and existing observations can lead to activate the reward, or 1 if it is feasible. Similarly, we will be interested in the max-marginal $\tilde p(r_{(-)}^{(t)}=0)$, i.e., whether it is feasible to find a configuration that avoids a negative reward.

At a high-level, planning proceeds as follows: Identify feasible desirable states (activating positive rewards and deactivating negative rewards), clamp their value to our desires by adding a unary potential to the factor graph, and then find the MAP configuration of the resulting graph. The MAP configuration contains the values of the action variables that are required to reach our goal of activating/deactivating a variable. We can also look at $S$ to see how the model ``imagines'' the evolution of the entities until they reach their goal state. Then we perform the actions found by the planner and repeat. We now explain each of these stages in more detail.

\paragraph{Potential feasibility analysis} First we run a feasibility analysis. To this end, a forward pass MPBP from time $0$ to time $T$ is performed. This provides a (coarse) approximation to the desired max-marginals for every variable. Because the SN graph is loopy, MPBP is not exact and the forward pass can be too optimistic, announcing the feasibility of states that are unfeasible\footnote{To illustrate the problem, consider the case in which it is feasible for an entity to move at time $t$ to position $A$ or position $B$ (but obviously not both) and then some reward is conditioned on that type of entity being in both positions: A single forward pass will not handle the entanglement properly and will incorrectly report that such reward is also feasible.}. Actual feasibility will be verified later, at the backtracking stage.

\paragraph{Choosing a positive reward goal state} We will choose the potentially feasible positive reward that happens soonest within our explored window, clamp its state to 1, and backtrack (see below) to find the set of actions that lead to it. If backtracking fails, we will repeat for the remaining potentially feasible positive rewards.

\paragraph{Avoiding negative rewards} Keeping the selected positive reward variable clamped to 1 (if it was found in the previous step), we now repeat the same procedure on the negative rewards. Among the negative rewards that have been found as potentially feasible to turn off, we clamp to zero as many negative rewards as we can find a jointly satisfying backtrack. If no positive reward was feasible, we backtrack from the earliest predicted negative reward.

\paragraph{Backtracking} This step is akin to Viterbi backtracking, a message passing backward pass that finds a satisfying configuration. Unlike the HMM for which the Viterbi algorithm was designed, our model is loopy, so a standard backward pass is not enough to find a satisfying configuration (although can help to find good candidates). We combine the standard backward pass with a depth-first search algorithm to find a satisfying configuration.

\section{Experiments}

\begin{figure*}[t]
\subfloat[][Mini Breakout Learning Rate]{\includegraphics[width=\columnwidth]{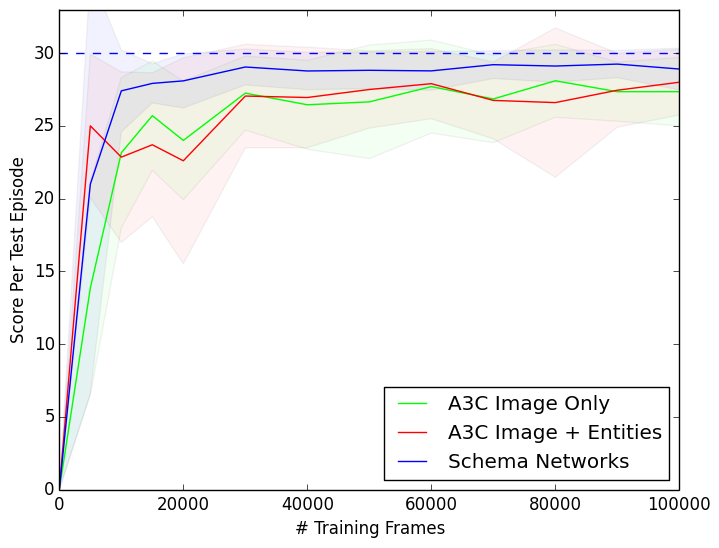}}
\subfloat[][Middle Wall Learning Rate]{\includegraphics[width=\columnwidth]{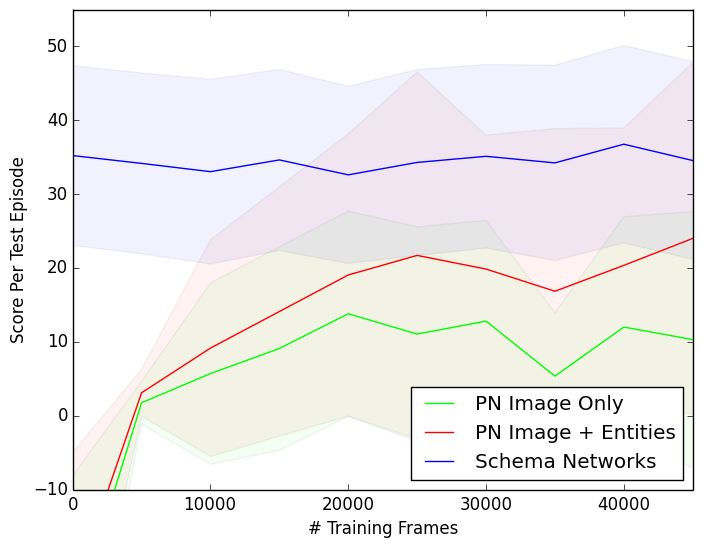}}
\caption{Comparison of learning rates. (a) Schema Networks and A3C were trained for 100k frames in Mini Breakout. Plot shows the average of 5 training attempts for Schema Networks and the best of 5 training attempts for A3C, which did not converge as reliably. (b) PNs and Schema Networks were pretrained on 100K frames of Standard Breakout, and then training continued on 45K additional frames of the Middle Wall variation. We show performance as a function of training frames for both models. Note that Schema Networks are ignoring all the additional training data, since all the required schemas were learned during pretraining. For Schema Networks, zero-shot transfer learning is happening.}
\vskip 0.15in
\label{fig:learning_rates}
\end{figure*}

\label{sec:experiments}
\label{experiments}

We compared the performance of Schema Networks, A3C, and PNs (Progressive Networks) on several variations of the game Breakout. The chosen variations all share similar dynamics, but the layouts change, requiring different policies to achieve high scores. A diverse set of concepts must be learned to correctly predict object movements and rewards. For example, when predicting why rewards occur, the model must disentangle possible causes to discover that reward depends on the color of a brick but is independent of the ball's velocity and position where it was hit. While these causal relationships are straightforward for humans to recover, we have yet to see any existing approach for learning a generative model that can recover all of these dynamics without supervision and transfer them effectively.

Schema Networks rely on an input of entity states instead of raw images, and we provided the same information to A3C and PNs by augmenting the three color channels of the image with 34 additional channels. Four of these channels indicated the shape to which each pixel belongs, including shapes for bricks, balls, and walls. Another 30 channels indicated the positions of parts of the paddle, where each part consisted of a single pixel. To reduce training time, we did not provide A3C and PN with part channels for objects other than the paddle, since these are not required to learn the dynamics or predict scores. Removing irrelevant inputs could only give A3C and PN an advantage, since the input to Schema Networks did not treat any object differently. Schema Networks were provided separate entities for each part (pixel) of each object, and each entity contained 53 attributes corresponding to the available part labels (21 for bricks, 30 for the paddle, 1 for walls, and 1 for the ball). Only one of these part attributes was active per entity. Schema Networks had to learn that some attributes, like parts of bricks, were irrelevant for prediction.

\subsection{Transfer Learning}
\label{exp:transfer_learning}

This experiment examines how effectively Schema Networks and PNs are able to learn a new Breakout variation after pretraining, which examines how well the two models can transfer existing knowledge to a new task. Fig.~\ref{fig:learning_rates}a shows the learning rates during 100k frames of training on Mini Breakout. In a second experiment, we pretrained on Large Breakout for 100k frames and continued training on the Middle Wall variation, shown in  Fig.~\ref{fig:breakout_variations}b. Fig.~\ref{fig:learning_rates}b shows that PNs require significant time to learn in this new environment, while Schema Networks do not learn anything new because the dynamics are the same.

\subsection{Zero-Shot Generalization}

\begin{table*}[tb]
\begin{center}
\caption{
\textbf{Zero-Shot Average Score per Episode}
Average of the 2 best out of 5 training attempts for A3C, and average of 5 training attempts for Schema Networks. A3C was trained on 200k frames of Standard Breakout (hence its zero-shot scores for Standard Breakout are unknown) while Schema Networks were trained on 100k frames of Mini Breakout. Episodes were limited to 2500 frames for all variations. In every case the average Schema Network scores are better than the best A3C scores by more than one standard deviation.
}

\vskip 0.15in

\begin{tabular}{
l
S[table-format=-1.2, table-figures-uncertainty=1]
S[table-format=-1.2, table-figures-uncertainty=1]
S[table-format=-1.2, table-figures-uncertainty=1]
S[table-format=-1.2, table-figures-uncertainty=1]
S[table-format=-1.2, table-figures-uncertainty=1]
S[table-format=-1.2, table-figures-uncertainty=1]
}
    \toprule[1pt]
    ~                     & {Standard Breakout} & {Offset Paddle} & {Middle Wall}   & {Random Target} & {Juggling} \\ \hline
    A3C Image Only        & \text{N/A}      & 0.60 \pm 20.05  & 9.55 \pm 17.44  & 6.83 \pm 5.02   & -39.35 \pm 14.57 \\
    A3C Image + Entities  & \text{N/A}      & 11.10 \pm 17.44 & 8.00 \pm 14.61  & 6.88 \pm 6.19 & -17.52 \pm 17.39 \\
    Schema Networks       & 36.33 \pm 6.17      & 41.42 \pm 6.29  & 35.22 \pm 12.23 & 21.38 \pm 5.02  & -0.11 \pm 0.34 \\
    \bottomrule[1pt]
\end{tabular}

\label{zeroshot}
\end{center}
\end{table*}

Many Breakout variations can be constructed that all involve the same dynamics. If a model correctly learns the dynamics from one variation, in theory the others could be played perfectly by planning using the learned model. Rather than comparing transfer with additional training using PNs, in these variations we can compare zero-shot generalization by training A3C only on Standard Breakout. Fig.~\ref{fig:breakout_variations}b-e shows some of these variations with the following modifications from the training environment:

\begin{itemize}
\item \textbf{Offset Paddle} (Fig.~\ref{fig:breakout_variations}d): The paddle is shifted upward by a few pixels.
\item \textbf{Middle Wall} (Fig.~\ref{fig:breakout_variations}b): A wall is placed in the middle of the screen, requiring the agent to aim around it to hit the bricks.
\item \textbf{Random Target} (Fig.~\ref{fig:breakout_variations}e): A group of bricks is destoyed when the ball hits any of them and then reappears in a new random position, requiring the agent to delibarately aim at the group.
\item \textbf{Juggling} (Fig.~\ref{fig:breakout_variations}f, enlarged from actual environment to see the balls): Without any bricks, three balls are launched in such a way that a perfect policy could juggle them without dropping any.
\end{itemize}

Table~\ref{zeroshot} shows the average scores per episode in each Breakout variation. These results show that A3C has failed to recognize the common dynamics and adapt its policy accordingly. This comes as no surprise, as the policy it has learned for Standard Breakout is no longer applicable in these variations. Simply adding an offset to the paddle is sufficient to confuse A3C, which has not learned the causal nature of controlling the paddle with actions and controlling the ball with the paddle. The Middle Wall and Random Target variations illustrate that Schema Networks are aiming to deliberately cause positive rewards from ball-brick collisions, while A3C struggles to adapt its policy accordingly. The Juggling variation is particularly challenging, since it is not clear which ball to respond to unless the model understands that the lowest downward-moving ball is the most imminent cause of a negative reward. By learning and transferring the correct causal dynamics, Schema Networks outperform A3C in all variations.

\begin{table}[tb]
\label{tab:zeroshot_half_neg}
\begin{center}

\caption{
\textbf{Average Score per Episode on Half Negative Bricks}
A3C and Schema Networks were trained on 200k frames of Random Negative Bricks, both to convergence. Testing episodes were limited to 1000 frames. Negative rewards are sometimes unavoidable, resulting in higher variance for all methods.
}
\vskip 0.15in

\begin{tabular}{
l
S[table-format=-1.2, table-figures-uncertainty=1]
}
    \toprule[1pt]
    ~                     & {Half Negative Bricks} \\ \hline
    A3C Image Only        & -3.80 \pm 7.55 \\
    A3C Image + Entities  & 1.55 \pm 6.41 \\
    Schema Networks       & 5.77 \pm 4.30 \\
    \bottomrule[1pt]
\end{tabular}

\end{center}
\end{table}

\subsection{Testing for Learned Causes}

To better evaluate whether these models are truly learning the causes of rewards, we designed one more zero-shot generalization experiment. We trained both Schema Networks and A3C on a Mini Breakout variation in which the color of a brick determines whether a positive or negative reward is received when it is destroyed. Six colors of bricks provide +1 reward, and two colors provide -1 reward. Negative bricks occurred in random positions 33\% of the time during training. Then during testing, the bricks were arranged into two halves, with all positive colored bricks on one half and negative colored bricks on the other (see Fig.~\ref{fig:breakout_variations}c). If the causes of rewards have been correctly learned, the agent should prefer to aim for the positive half whenever possible. As Table 1 shows, Schema Networks have correctly learned from random arrangements which brick colors cause which rewards, preferring to aim for the positive half during testing, while A3C demonstrates no preference for one half or the other, achieving an average score near zero.

\section{Discussion and Conclusion}

In this work we have demonstrated the promise of Schema Networks with strong performance on a suite of Breakout variations. Instead of learning policies to maximize rewards, the learning objective for Schema Networks is designed to \textit{understand causality} within these environments. The fact that Schema Networks are able to achieve rewards more efficiently than state-of-the-art model-free methods like A3C is all the more notable, since high scores are a byproduct of learning an accurate model of the game.

The success of Schema Networks is derived in part from the entity-based representation of state. Our results suggest that providing Deep RL models like A3C with such a representation as input can improve both training efficiency and generalization. This finding corroborates recent attempts \cite{usunier2016episodic, garnelo2016towards, chang2016compositional, battaglia2016interaction} to incorporate object and relational structure into neural network-based models.

The environments considered in this work are conceptually diverse but also simplified in a number of ways with respect to the real world: states, actions, and rewards are all discretized as binary random variables; the dynamics of the environments are deterministic; and there is no uncertainty in the observed entity states. In future work we plan to address each of these limitations, adapting Schema Networks to continuous, stochastic domains.

Schema Networks have shown promise toward multi-task transfer where Deep RL struggles. This transfer is enabled by explicit causal structures, which in turn allow for planning in novel tasks. As progress in RL and planning continues, robust generalization from limited experience will be vital for future intelligent systems.

\section*{Acknowledgements}

Special thanks to Eric Purdy and Ramki Gummadi for useful insights and discussions during the preparation of this work.


\bibliography{icml2017-schemas}
\bibliographystyle{icml2017}

\appendix

\section{Breakout playing visualizations}

See \url{https://vimeopro.com/user45297729/schema-networks} for visualizations of Schema Networks playing different variations of Breakout after training only on basic Breakout.

\begin{figure}[!h]
  \centering
  \subfloat[A3C]{\includegraphics[width=0.475\columnwidth]{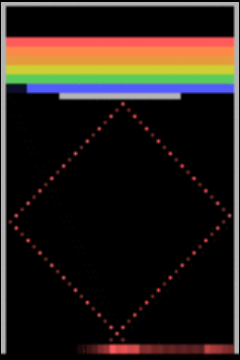}}
  \hfill
  \subfloat[Schema Networks]{\includegraphics[width=0.5\columnwidth]{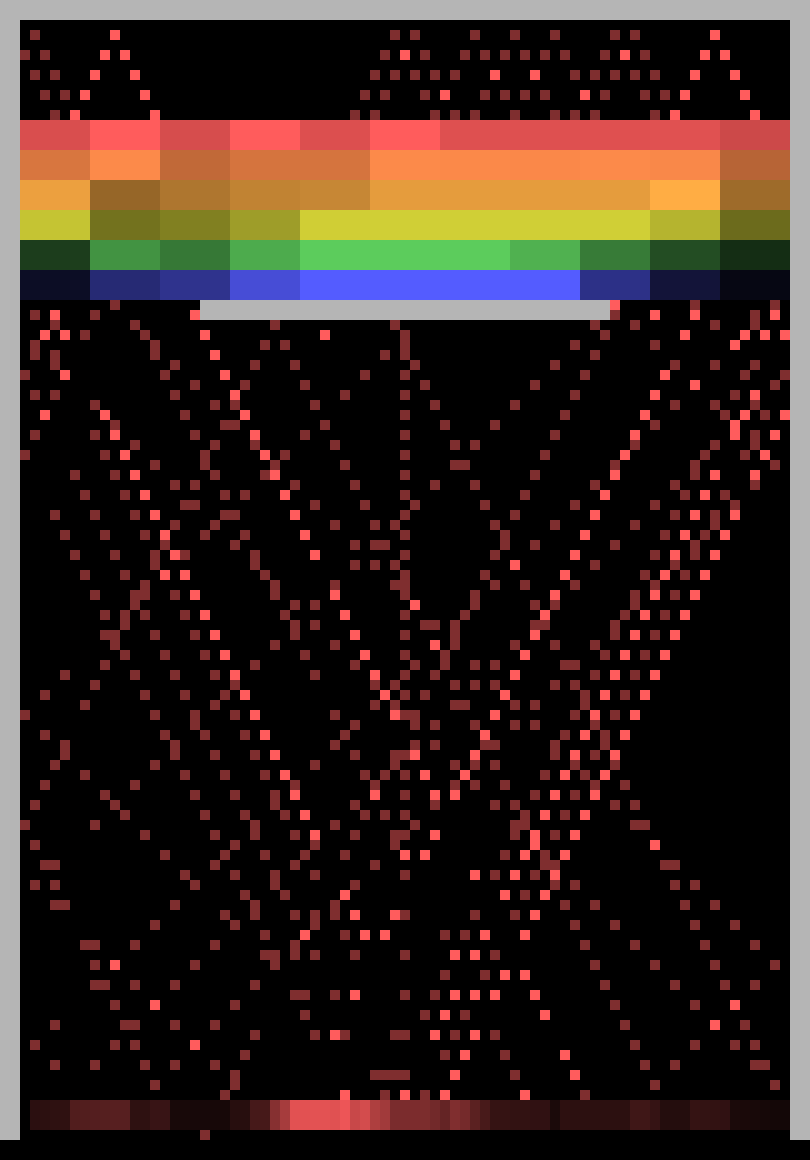}}
  \caption{Screen average during the course of gameplay in the Midwall variation when using A3C and Schema Networks. SNs are able to purposefully avoid the middle wall most of the time, whereas A3C struggles to score any points.}
\label{fig:comparison}
\end{figure}

Figure \ref{fig:comparison} shows typical gameplay for one variation.

\pagebreak
\section{LP-based Greedy Schema Learning}

The details of the LP-based learning of Section 4.2 are provided here.

\renewcommand{\algorithmicrequire}{\textbf{Input:}}
\renewcommand{\algorithmicensure}{\textbf{Output:}}

\begin{algorithm}[htb]
\caption{LP-based greedy schema learning}\label{alg:schemalearning}
\begin{algorithmic}[1]
\REQUIRE Input vectors $\{x_n\}$ for which $f_W(x_n) = 0$ (the current schema network predicts 0), and the corresponding output scalars ${y_n}$
\STATE \textbf{Find a cluster of input samples} that can be solved with a single (relaxed) schema while keeping perfect precision (no false alarms). Select an input sample and put it in the set ``solved'', then solve the LP
\begin{align*}
\min_{w\in [0,1]^{D}} & \sum_{n: y_n=1} (1-x_n)w\\
\text{s.t. } & (1-x_n)w > 1 \quad \forall_{n: y_n=0}\\
& (1-x_n)w = 0 \quad \forall_{n \in \text{solved}}\\
\end{align*}
\STATE \textbf{Simplify the resulting schema}. Put all the input samples for which $(1-x_n)w = 0$  in the set ``solved''. Simplify the just found schema $w$ by making it as sparse as possible while keeping the same precision and recall:
\begin{align*}
\min_{w\in [0,1]^{D}} & w^T\vec{1}\\
\text{s.t. } & (1-x_n)w > 1 \quad \forall_{n: y_n=0}\\
& (1-x_n)w = 0 \quad \forall_{n \in \text{solved}}\\
\end{align*}
\STATE \textbf{Binarize the schema}. In practice, the found $w$ is binary most of the time. If it is not, repeat the previous minimization using binary programming, but optimize only over the elements of $w$ that were found to be non-zero. Keep the rest clamped to zero.

\ENSURE New schema $w$ to add to the network
\end{algorithmic}
\end{algorithm}

\end{document}